\documentclass[runningheads]{llncs}
\usepackage{graphicx}
\usepackage{amsmath,amssymb} 
\usepackage{color}
\definecolor{ForestGreen}{RGB}{34,139,34}

\usepackage{listings}
\usepackage{xcolor}
\usepackage{tikz}
\usepackage{comment}
\usepackage{amsmath,amssymb} 
\usepackage{color}
\usepackage{booktabs} 

\usepackage{pifont}

\usepackage{makecell}
\definecolor{codegreen}{rgb}{0,0.6,0}
\definecolor{codegray}{rgb}{0.5,0.5,0.5}
\definecolor{codepurple}{rgb}{0.58,0,0.82}
\definecolor{backcolour}{rgb}{0.95,0.95,0.92}

\lstdefinestyle{mystyle}{
    backgroundcolor=\color{backcolour},   
    commentstyle=\color{codegreen},
    keywordstyle=\color{magenta},
    numberstyle=\tiny\color{codegray},
    stringstyle=\color{codepurple},
    basicstyle=\ttfamily\footnotesize,
    breakatwhitespace=false,         
    breaklines=true,                 
    captionpos=b,                    
    keepspaces=true,                 
    numbers=left,                    
    numbersep=5pt,                  
    showspaces=false,                
    showstringspaces=false,
    showtabs=false,                  
    tabsize=2
}

\usepackage{graphicx}
\usepackage{comment}
\usepackage{amsmath,amssymb} 
\usepackage{color}
\usepackage{caption}
\usepackage{subcaption}
\usepackage[width=122mm,left=12mm,paperwidth=146mm,height=193mm,top=12mm,paperheight=217mm]{geometry}

\usepackage{amssymb}
\usepackage{pifont}
\usepackage{floatrow}
\usepackage{floatrow}
\usepackage{setspace}
\usepackage{ulem}
\usepackage{xstring}
\usepackage{footnote}
\usepackage{color}

\usepackage{multirow}
\makeatother
\usepackage[bookmarks=false]{hyperref}
\begin{document}
\title{PreTraM: Self-Supervised Pre-training via Connecting Trajectory and Map} 



\titlerunning{PreTraM}
%

%
\author{Chenfeng Xu\inst{1}\thanks{Equal contribution} \and
Tian Li\inst{3}* \and Chen Tang\inst{1}  \and Lingfeng Sun\inst{1} \and Kurt Keutzer\inst{1} \and Masayoshi Tomizuka\inst{1} \and Alireza Fathi \inst{2} \and Wei Zhan \inst{1}}
%
\authorrunning{C. Xu et al.}
%

\institute{University of California, Berkeley \and Google Research \and University of California San Diego
\email{\{xuchenfeng,chen\_tang,lingfengsun,wzhan\}@berkeley.edu},\\
\email{tianli@ucsd,edu}, \\
 \email{alirezafathi@google.com}\\
}
\maketitle    
%
\begin{abstract}
Deep learning has recently achieved significant progress in trajectory forecasting. However, the scarcity of trajectory data inhibits the data-hungry deep-learning models from learning good representations. 
While mature representation learning methods exist in computer vision and natural language processing, these pre-training methods require large-scale data. It is hard to replicate these approaches in trajectory forecasting due to the lack of adequate trajectory data (e.g., 34K samples in the nuScenes dataset).
To work around the scarcity of trajectory data, we resort to another data modality closely related to trajectories—HD-maps, which is abundantly provided in existing datasets. In this paper, we propose \textit{PreTraM}, a self-supervised \textbf{pre}-training scheme via connecting \textbf{tra}jectories and \textbf{m}aps for trajectory forecasting. Specifically, PreTraM consists of two parts: 1) Trajectory-Map Contrastive Learning, where we project trajectories and maps to a shared embedding space with cross-modal contrastive learning, and 2) Map Contrastive Learning, where we enhance map representation with contrastive learning on large quantities of HD-maps. On top of popular baselines such as AgentFormer and Trajectron++, PreTraM boosts their performance by 5.5\% and 6.9\% relatively in FDE-10 on the challenging nuScenes dataset. We show that PreTraM improves data efficiency and scales well with model size.

\end{abstract}

%

\section{Introduction}
\label{intro}
Trajectory forecasting is a challenging task in autonomous driving, which aims at predicting the future trajectory conditioned on past trajectories and surrounding scenes. Current deep learning models have dominated trajectory forecasting by data-driven supervised learning. However, both the collection and the annotation of trajectory data are extremely difficult and costly. Trajectory data is collected by self-driving vehicles with sophisticated sensor systems.
Then annotators need to label the objects, associate their positions, generate and smoothen trajectories. This complex procedure limits the scale of the data. For example, the popular open-sourced ``large-scale" trajectory forecasting dataset nuScenes \cite{caesar2020nuScenes} has only 34K samples, even much less than that of the elementary small-scale image dataset MNIST (60K samples) \cite{deng2012mnist}. The scarcity of trajectory data prohibits the models from learning good trajectory representation, which restrains their performance when trained with such a small amount of data. 


In the Natural Language Processing (NLP) and computer vision (CV) communities, it was found effective to use self-supervised pre-training on vast unlabeled datasets to learn language/visual representations. The classic methods, such as autoregressive language modeling \cite{brown2020language}, masked autoencoding \cite{devlin2019bert}, and contrastive learning~\cite{he2020momentum,chen2020simple}, are conceptually simple, but require billions of training data. Although recent results from CLIP \cite{radford2021learning} show that cross-modal contrastive learning requires much fewer pre-training data (4x fewer), the amount of data used is still far more than available trajectory data. 
Unlike NLP and CV, where large-scale unlabeled datasets exist for pre-training, the bottleneck for scaling trajectory datasets lies in data collection and annotation. 
It makes the trajectory prediction field hard to benefit from common pre-training schemes. 
As a result, to the best of our knowledge, few efforts in trajectory forecasting have explored pre-training.

\begin{figure}[!t]
	\begin{center}
		\includegraphics[width=0.95\linewidth]{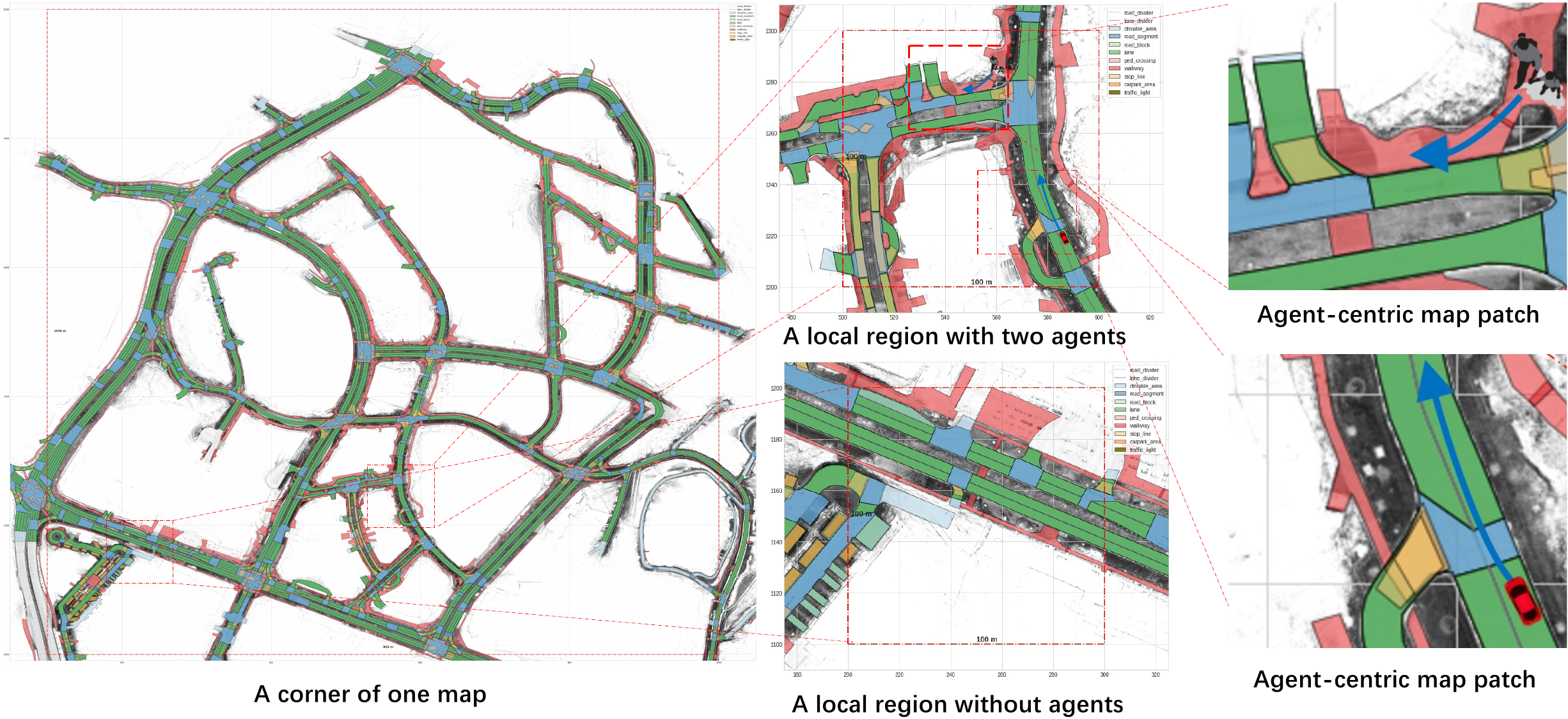}
	\end{center}
	\caption{We have two key observation about maps and trajectories: 1) As shown in the rightmost column, vehicles usually move in drivable areas and pedestrians usually move along sidewalks. And the relationship learn from scenes like the one in the top middle can generalize to unseen scenes in the bottom middle. (Please zoom in for better view.) 2) Agent-centric map patches are taken from a local region of the map, which is just a tiny part of the whole map. }
	\label{fig:intro1}
\end{figure}

To work around the scarcity of trajectories, we resort to another modality of data that is closely related to trajectories—HD-maps. In fact, we observe two important facts about maps:
\begin{itemize}
\item An agent's trajectory is correlated to the map around it \cite{vectornet,scenetransformer}. A representative example is that the shape of trajectory usually follows the topology of the HD-map. As shown in the rightmost column of Figure~\ref{fig:intro1}, vehicles usually move in drivable areas, and pedestrians usually move along sidewalks. More importantly, the relationships between trajectory and map can be generalized to other scenes. 
For example, in the middle of Figure~\ref{fig:intro1}, the model learns from the upper scene that the moving car should \textit{follow the boundary of the road}. By capturing this relationship, the model knows that a car in the unseen bottom scene should also \textit{follow the boundary of the road}.
If the model can capture these trajectory-map relationships, it is viable to learn useful trajectory representations without large amounts of annotated trajectories.

\item Existing works in trajectory forecasting only take advantage of the agent-centric map patches, the local regions containing trajectories of at least one annotated agent, but significantly under-utilize other parts of the maps, which cover much larger areas. As shown in Figure~\ref{fig:intro1}, agent-centric map patches are tiny compared with the leftmost global map, leaving most parts of the map unexploited. 

\end{itemize}





Based on the above observations, we propose PreTraM, a self-supervised \textbf{pre}-training scheme via connecting \textbf{tra}jectories and \textbf{m}aps for trajectory forecasting. Specifically, we jointly pre-train the trajectory and map encoders of a trajectory prediction model in two ways: 
1) \textit{Trajectory-Map Contrastive Learning (TMCL)}: 
Inspired by CLIP \cite{radford2021learning}, we constrast trajectories with corresponding map patches to enforce the model to capture their relationship.
2) \textit{Map Contrastive Learning (MCL)}: We train a stronger map encoder with contrastive learning on large quantities of trajectory-decoupled map patches, which outnumber the agent-centric ones by 782x. In short, PreTraM is a synergy of TMCL and MCL: TMCL benefits trajectory representation by understanding trajectory-map relationship; MCL further enhances that understanding by improving map representation.



Our method boosts the performance of  
a variety of popular prediction models like AgentFormer \cite{yuan2021agent} and Trajectron++ \cite{salzmann2020trajectron++} by 5.5\% and 6.9\% relatively in FDE-10 on the nuScenes dataset \cite{caesar2020nuScenes}. 
More importantly, we find that PreTraM is able to achieve larger performance gain when data gets fewer. 
Impressively, using only 70 \% of the trajectory data, PreTraM on top of AgentFormer show superior performance compared with the baseline trained on 100 \% trajectory data. 
This demonstrates the proposed pre-training scheme brings strong data efficiency.
Furthermore, we apply PreTraM to larger versions of AgentFormer and observe that it consistently improves prediction accuracy when the model scales up. We also conduct sufficient ablation studies and provide analysis to demonstrate the effectiveness of PreTraM and shed light on how it works. 


In summary, our key contributions are as follows:
\begin{itemize}
    \item We propose PreTraM, a novel self-supervised pre-training scheme for trajectory forecasting by connecting trajectories and maps, which consists of trajectory-map contrastive learning and map contrastive learning. 
    
    \item We show with experiments that PreTraM achieves up to 6.9 \% relative improvement in FDE-10 upon popular baselines.
    
    \item PreTraM enhances the data efficiency of prediction models, using 70\% training data but beating baseline with 100\% training data, and generalizes to models with larger scales. 
    
    \item Through ablation studies and analysis, we demonstrate the efficacy of TMCL and MCL respectively, and shed light on how PreTraM works.
\end{itemize}

\section{Background}
\subsection{Problem Formulation of Trajectory Forecasting}
In trajectory forecasting, we aim to predict the future trajectories of multiple target agents in a scene. Typically, a set of history states $x$ for all agents and the surrounding HD-map patches $M$ are input to the model $f_\omega$ and the model predicts the future trajectories of each agent $y = f_\omega(x,M)$.

The HD-map contains rich semantic information (e.g., drivable area, stop line, and traffic light)~\cite{caesar2020nuScenes}. In this work, we employ rasterized top-down semantic images around each of the agents as the input HD-map patches $M$, i.e., $M=\{m_i\}_{i\in\{1,...,A\}}, m_i\in \mathbb{R}^{C\times C\times 3}$, where $C$ is the context size and 3 denotes the RGB channels. Note that each color has its specific semantic meaning in HD-maps.

As for the history states, denoting the number of agents in the scene as $A$, and the history time span as $T$, then $x = s_{1,...,A}^{(-T:0)}\in \mathbb{R}^{T\times A\times D}$, where $s_i$ is the history states of agent $i$, and 0 denotes the current timestamp. $D$ is the dimension of features that generally contain the agent's 2D or 3D coordinates, as well as other information such as its heading and its speed. 

\subsection{Contrastive Learning}
Contrastive learning is a powerful method for self-supervised representation learning that was made popular by~\cite{he2020momentum,chen2020improved,chen2020simple,chen2020big}. 
Using instance discrimination as the pretext task, they pull the semantically-close neighbors together and push away non-neighbors~\cite{gao2021simcse}. 
For example, in SimCLR~\cite{chen2020simple}, given a mini-batch of inputs, each input $x_i$ is transformed into a positive sample $x_i^+$. Let $h_i, h_i^+$ denote the hidden representation of $x_i, x_i^+$. Then on a mini-batch of $N$ pairs of $(x_i, x_i^+)$, it adopts the InfoNCE loss~\cite{van2018representation} as its training objective.


In particular, we are interested in one specific work that explored contrastive learning in NLP: SimCSE~\cite{gao2021simcse}. Instead of using word replacement or deletion as augmentation, it uses different dropout masks in the model as the minimal augmentation for the positive samples. This simple approach turns out to be very effective in that it fully preserves the semantic of the text, compared with other augmentation operations. To preserve the semantic of HD-map, we also adopt dropout for the positive samples in map contrastive learning. 

More recently, CLIP~\cite{radford2021learning} demonstrated the power of cross-modal contrastive learning conditioned on huge amounts of data. It collects paired images and captions from the Internet and asks the model to pair an image with the corresponding text, using large batches. For a mini-batch of $N$ pairs of images $I_i$ and texts $T_i$, denoting their hidden representations as $(h_i^I, h_i^T)$, it applies cross-entropy loss on the $N\times N$ similarity matrix over all pairs of images and texts, stated as follows:

\begin{equation}
    l_i = -\log\frac{e^{\text{sim}(h_i^I,  h_i^T)/\tau}}{\sum_{j=0}^Ne^{\text{sim}(h_i^I, h_j^T)/\tau}}
\end{equation}
where $\text{sim}(\cdot)$ is a measurement of similarity, typically the cosine similarity, and $\tau$ is the temperature parameter.
Note that it can be seen as the InfoNCE loss using the corresponding text as the positive sample of an image.

Intuitively, using natural language as supervision of images, CLIP puts image and text in a shared embedding space. Besides, equation (1) enforces similarity between the correct pair of images and text, and thus learns the pattern of image-text relationship. Following this intuition, we design a trajectory-map contrastive learning objective to capture the relationship between them.



\begin{figure}[!t]
    \centering
    \includegraphics[width=0.9\textwidth]{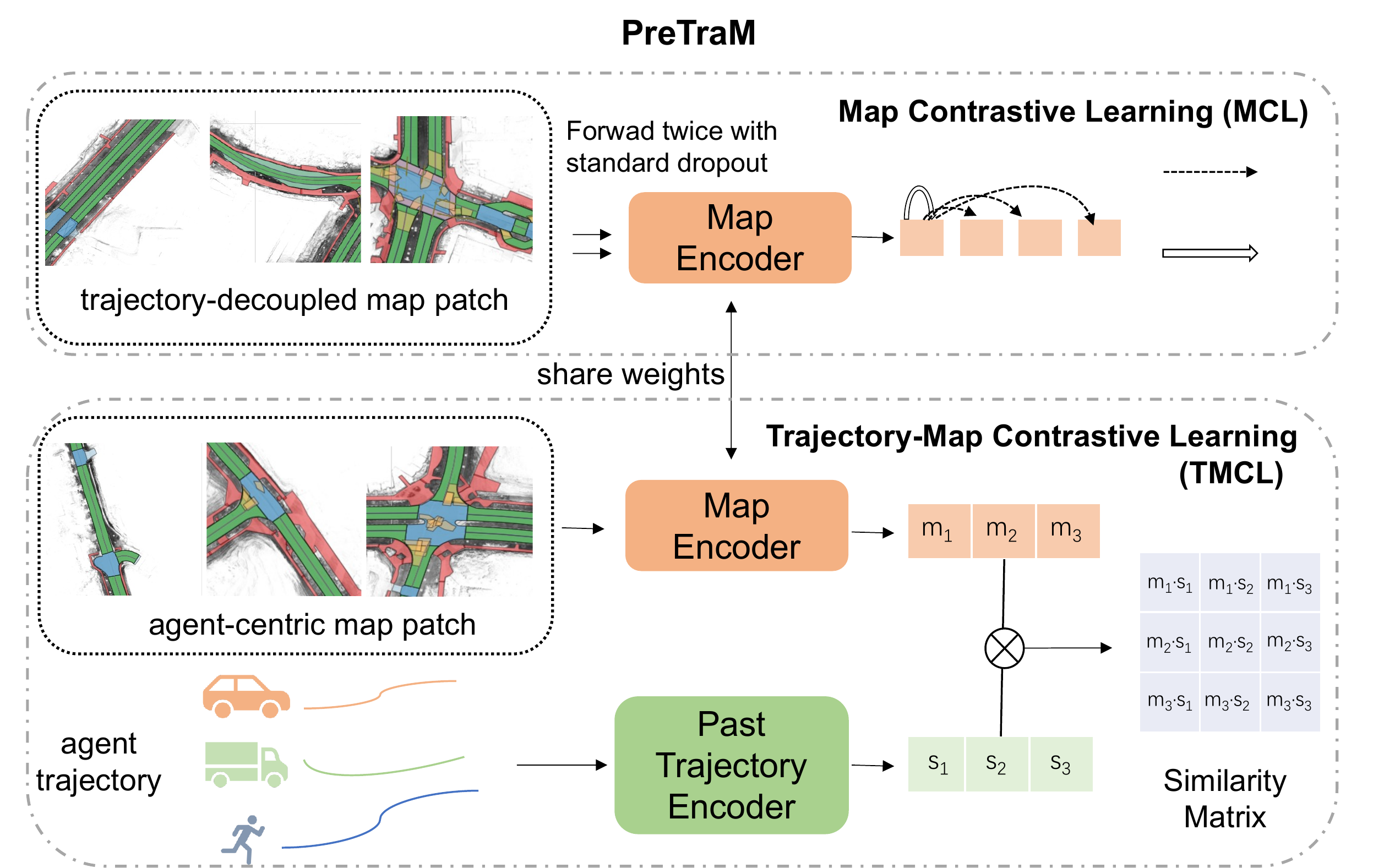}
    \caption{Top: Map Contrastive Learning (MCL). On the contrary to agent-centric map patches, trajectory-decoupled ones do not necessarily contain agent trajectories. During training, we randomly crop those patches from the whole map around positions on the road. Bottom: Trajectory-Map Contrastive Learning (TMCL). }
    \label{fig:method}
\end{figure}

\section{Method}
\label{method}

We propose a novel self-supervised \textbf{pre}-training scheme by connecting \textbf{tra}jectory and \textbf{m}ap (PreTraM) to enhance the trajectory and map representations when there are small-scale trajectory data, but large-scale map data. We jointly pre-train a trajectory encoder and a map encoder to obtain good trajectory representation by encoding the trajectory-map relationship into the representation.

As illustrated in Figure \ref{fig:method}, the proposed PreTraM is composed of two parts: 1) A simple trajectory-map contrastive learning (TMCL) that is conducted between map encoder and trajectory encoder, using limited trajectory and the paired map patches. 2) A simple map contrastive learning (MCL) that is conducted on map encoder using large batch size on trajectory-decoupled map patches, where there are not necessarily agent trajectories. After pre-training, we load the pre-trained weights and finetune under the prediction objective with the same training schedules as the original models.

\subsection{Trajectory-Map Contrastive Learning (TMCL)}
\label{sec:TMCL}
We propose to use a cross-modal contrastive learning method that facilitates both trajectory encoder and map encoder. Specifically, given a mini-batch of scenes, for all the input history states $x$, we split them into single agent trajectories and treat them \textit{independently}, i.e., $S=\{s_i|s_i\in x, \forall x\in B\}$, where $B$ denotes the mini-batch. For each agent, we crop an agent-centric map patch around its current position from the HD-map, and then we have $N_{\text{traj}}=|S|$ pairs of correlated trajectories and HD-map patches $(s_i, m_i)$. We also rotate the map with respect to the orientation of the agent following the common practice~\cite{yuan2021agent,salzmann2020trajectron++}. 
The model is required to match each trajectory $s_i$ with the paired map patch $m_i$ among all the map patches in the mini-batch and vice versa. As shown in bottom of Figure \ref{fig:method}, we input the trajectories and maps into the corresponding encoders to obtain the features $\{h^{\text{traj}}_i\}, \{h^{\text{map}}_i\}$. Then we compute a similarity matrix across all pairs of trajectories and maps in the mini-batch. 
Note that we apply a linear projection layer~\cite{radford2021learning} after the map encoder and the trajectory encoder for the hidden representations $h$ but omit it above for the sake of simplicity. It is the same case in section~\ref{sec:MCL} for MCL. 
Finally, we optimize a symmetric cross-entropy loss over these similarity scores as follows~\cite{radford2021learning}:
\begin{equation}
    l_i^{\text{TMCL}} = -\log \frac{e^{\text{sim}(h^{\text{traj}}_i,h^{\text{map}}_i)/\tau_{\text{traj}}}}{\sum_{j=1}^{N_{\text{traj}}}e^{\text{sim}(h^{\text{traj}}_i,h^{\text{map}}_j)/\tau_{\text{traj}}}}
\end{equation}

Through this objective, the similarities of the correct pairs of the trajectories and maps are maximized and those of the other pairs are minimized. It results in a shared embedding space of trajectories and maps. We find that a prediction model, which fuses map and trajectories to make prediction, benefit from such a shared embedding space. It agrees with the finding in~\cite{ALBEF} that models for vision-language tasks benefit from an aligned embedding space for the visual and language inputs. The TMCL objective teaches the model to encode the relationship between maps and trajectories into the representation. By capturing the relationship, the trajectory embedding contains the information of the underlying map conditioned on the input trajectory, which implies the geometric and routing information of the future trajectories for the predictor. 

\subsection{Map Contrastive Learning (MCL)}
\label{sec:MCL}

To further facilitate learning the trajectory-map relationship, we learn a general map representation by map contrastive learning. 
At each training iteration, we randomly crop $N_{\text{map}}$ map patches from a random subset of HD-Maps in the dataset.
Note that $N_{\text{map}}$ is much greater than the agent number $A$.

In addition to using a large number of map patches, the key ingredient to get MCL to work effectively is using the exact identical instance as its positive sample, i.e., $m_i^+=m_i$, and apply dropout in the map encoder~\cite{gao2021simcse}. Denote the hidden representation $h_i^z = g_\theta(m_i,z)$ where $g_\theta$ is the map encoder and $z$ is a random mask for dropout. 
As shown in the top part of Figure \ref{fig:method}, we feed the the same map patch to the encoder in two independent forward passes with different dropout masks $z, z^\prime$, which gives two representation $h_i^z, h_i^{z^\prime}$ for each $m_i$. Thanks to dropout, $h_i^z$ and $h_i^{z^\prime}$ are different, but still encode the same topology and semantic. In contrast, regular augmentation operations in CV such as random rotation, flip, gaussian noise or color jitter do not work here. Gaussian noise and color jitter are prone to change the semantics of HD-maps, while flip and rotation change the topology structure of HD-maps. Instead, dropout serves as a minimal augmentation for the positive sample and turns out to be effective through experiments. Formally, the training objective of MCL is:
\begin{equation}
    l_i^{\text{MCL}} = -\log\frac{e^{\text{sim}(h_i^{z_i},h_i^{z_i^\prime})/\tau_{\text{map}}}}{\sum_{j=1}^{N_{\text{map}}} e^{\text{sim}(h_i^{z_i},h_j^{z_j^\prime})/\tau_{\text{map}}}}
\end{equation}

It is worth noting that MCL is a novel use of HD-maps not only because we make use of every piece of the HD-map, but also because we design a customized training objective to make better use of HD-maps. 




\subsection{Training Objective}
The overall pre-training scheme is the joint of trajectory-map contrastive learning (TMCL) and map contrastive learning (MCL). The overall objective function combines their objectives, given by: 
\begin{equation}
    \mathcal{L} = \sum_{i=1}^{N_\text{traj}} l_i^{\text{TMCL}} + \lambda\sum_{i=1}^{N_\text{map}} l_i^{\text{MCL}} 
\end{equation}

\section{Experiments}
\subsection{Dataset and implementation details}
\textbf{Dataset.} 
nuScenes is a recent large-scale autonomous driving dataset collected from Boston and Singapore. It consists of 1000 driving scenes with each scene annotated at 2Hz, and the driving routes are carefully chosen to capture challenging scenarios. The nuScenes dataset provides HD semantic maps from Boston Seaport together with Singapore’s One North, Queenstown and Holland Village districts, with 11 semantic classes. It is split into 700 scenes for training, 150 scenes for validation, and 150 scenes for testing. 

Our main experiments follow the split used in AgentFormer~\cite{yuan2021agent}, in which the original training set is split into two parts: 500 scenes for training, and 200 scenes for validation. The original validation set is used for testing our model.

\subsubsection{Baseline.} We performed experiments with Pre-TraM on two models, AgentFormer \cite{yuan2021agent} and Trajectron++ \cite{salzmann2020trajectron++}. Both of them are CVAE models including a past trajectory encoder, a map encoder, a future trajectory encoder, and a future trajectory decoder.
We reproduced AgentFormer to support parallel training. Compared with the original code, our reproduced code trains 17.1x faster than the official code (4.5 hours vs. 77 hours on one V100 GPU.), and its performance is competitive—0.029 better than the official implementation on ADE-5. Note that AgentFormer separately trains DLow~\cite{yuan2020dlow} for better sampling. We did not reproduce this part since we focus on representation learning and want a precise quantitative evaluation on the benefit of PreTraM to the model itself. Plus, Trajectron++ does not use DLow while applicable. We want to keep the setting consistent, so that it is meaningful to compare the performance gains between different models. As for Trajectron++ \cite{salzmann2020trajectron++}, we use their official implementation but re-train it using the data split in AgentFormer to ensure fair comparison. 
In the following sections, we denote AgentFormer/Trajectron++ pre-trained with PreTraM as PreTraM-AgentFormer/PreTraM-Trajectron++, or PreTraM when the model is clear in the context.

\subsubsection{Pre-training and finetuning. }
\label{implementation}
Our pre-training is applied to the \textit{past} trajectory encoder and map encoder. To train TMCL, we pair the historical trajectories of last 2s and map patches of context size $100 \times 100$. We randomly rotate the trajectories and maps simultaneously for data augmentation. For MCL, we collect the trajectory-decoupled map patches dynamically at training. For each instance in the mini-batch, we crop 120 map patches centered at random positions along the road in the HD-map. 
We pre-train the encoders with the PreTraM objective function for 20 epochs using batch size 32 (which means 3440 map patches for MCL in one iteration). 
Throughout the pre-training phase, we use 28.8M map patches to train our map encoder, which is 782x more than agent-centric map patches. 
The overall pre-training phase is fast—only \textit{30 minutes} on one V100 GPU for AgentFormer.

Recall that we use dropout for positive samples in MCL. In shallow map encoders, such as Map-CNN used in AgentFormer and Trajectron++, we place the dropout at post-activation of each convolution. For relatively deeper map encoder such as ResNet family, we place two dropout masks on each residual block. The mask ratio of dropout is default as $p=0.1$. 

At finetuning phase, we use the same training recipes as AgentFormer and Trajectron++. The prediction horizon is 6 seconds and we use the ground-truth future trajectories to supervised the training.

\subsubsection{Metric.} We use the common trajectory prediction metrics Average Displacement Error (ADE) and Final Displacement Error (FDE). We follow previous works \cite{yuan2021agent,salzmann2020trajectron++} to sample $k$ trajectories during inference and pick the minimum of the error, denoting as ADE-$k$ and FDE-$k$. Apart from the sampling based metrics above, we also use a deterministic metric meanFDE, which is the FDE of the trajectory that the model deems as the most likely. 

We also leverage the metrics including Kernel Density Estimate-based Negative Log Likelihood (KDE NLL)~\cite{salzmann2020trajectron++} and boundary violation rate. The former measures the NLL of the ground truth trajectory under a distribution created by fitting a kernel density estimate on trajectory samples, which shows the likelihood of the ground truth trajectory given the sampled trajectory predictions.
The latter is the ratio of the predicted trajectories that hit road boundaries.

\begin{table}[!t]

\caption{Comparison experiments based on AgentFormer \cite{yuan2021agent} and Trajectron++ \cite{salzmann2020trajectron++}. Note that the reported AgentFormer is removed of DLow. The AgentFormer* denotes our reproduced implementation. \textit{Lower} number is better.}
\centering
\begin{tabular}{ c c c c c }
 \hline
Method & ADE-5 & FDE-5 & ADE-10 & FDE-10\\
 \hline
MTP \cite{cui2019multimodal} &2.93& - & - & - \\
AgentFormer \cite{yuan2021agent} &2.517 & 5.459 & 1.852 & 3.869 \\
MultiPath \cite{chai2019multipath} & 2.32 & - & 1.96& - \\
DLow-AF \cite{yuan2020dlow} & 2.11& 4.70 & 1.78 & 3.58 \\
DSF-AF \cite{ma2020diverse} & 2.06 & 4.67 &1.66 & 3.71 \\
CoverNet \cite{phan2020covernet}  & 1.96 & - & 1.48& - \\
\hline
AgentFormer* & 2.488 & 5.420 & 1.893&3.902 \\
PreTraM-AgentFormer* &\textbf{2.391}\color{ForestGreen}{(-0.097)} &\textbf{5.177}\color{ForestGreen}{(-0.243)} & \textbf{1.796}\color{ForestGreen}{(-0.097)} & \textbf{3.687}\color{ForestGreen}{(-0.215)} \\
\hline
Trajectron++ \cite{salzmann2020trajectron++} & 1.772 & 4.150 & 1.405 & 3.221\\
PreTraM-Trajectron++ & \textbf{1.698}\color{ForestGreen}{(-0.074)}& \textbf{3.963}\color{ForestGreen}{(-0.197)}&  \textbf{1.348}\color{ForestGreen}{(-0.057)} & \textbf{3.040}\color{ForestGreen}{(-0.181)} \\
\hline
\end{tabular}
\label{tab:comparison_exp}
\end{table}

\begin{table}[!t]
\centering

\caption{Experimental evaluation on meanFDE, KDE NLL, and Boundary violation rate (B. Viol.) provided by Trajectron++ \cite{salzmann2020trajectron++}. \textit{Lower} number is better.}
\begin{tabular}{ c c c c}
 \hline
Method & meanFDE  & KDE NLL & B. Viol. (\%)\\
 \hline
Trajectron++ & 8.242 & 2.487 & 23.7\\
PreTraM-Trajectron++ & \textbf{8.212}\color{ForestGreen}{(-0.030)} &\textbf{2.380}\color{ForestGreen}{(-0.107)} &\textbf{21.9}\color{ForestGreen}{(-1.8)}\\

\hline
\end{tabular}
\label{tab:other metric}
\end{table}

\subsection{Comparison experiments}
The results compared with the baselines and the other prior-arts are shown in Table~\ref{tab:comparison_exp}. Observe that using PreTraM improves the performance by 0.097 (\textit{resp. 0.074}) ADE-5,  0.243 (\textit{resp. 0.197}) FDE-5,  0.097 (\textit{resp. 0.057}) ADE-10,  0.215 (\textit{resp. 0.181}) FDE-10, on top of AgentFormer (\textit{resp.} Trajectron++). This is up to \textit{4.1\%} relative improvement on ADE-5 and \textit{6.9\%} relative improvement on FDE-10. Remarkably, we achieve this improvement with a simple pre-training scheme. PreTraM does not rely on long pre-training epochs or huge quantities of external data as said in section \ref{implementation}. The extra HD-map data we use during pre-training is also provided by the dataset. It can be easily applied to almost arbitrary prediction model that fuses HD-map and trajectory to boost its performance. In conclusion, these results demonstrate that PreTraM indeed faciltates the models in representation learning. Note that the performance is evaluated on the validation set of nuScenes dataset, following AgentFormer \cite{yuan2021agent}.

We also evaluate the results on the metrics provided by Trajectron++ to show the advantage of PreTraM. PreTraM-Trajectron++ improves baseline by 0.107 KDE NLL and 1.8\% boundary violation rate (Table \ref{tab:other metric}). The improvements on these two metrics show that our pre-training scheme not only improves prediction accuracy, but also improves stability and safety.

\begin{figure}[!t]
	\begin{center}
		\includegraphics[width=\linewidth]{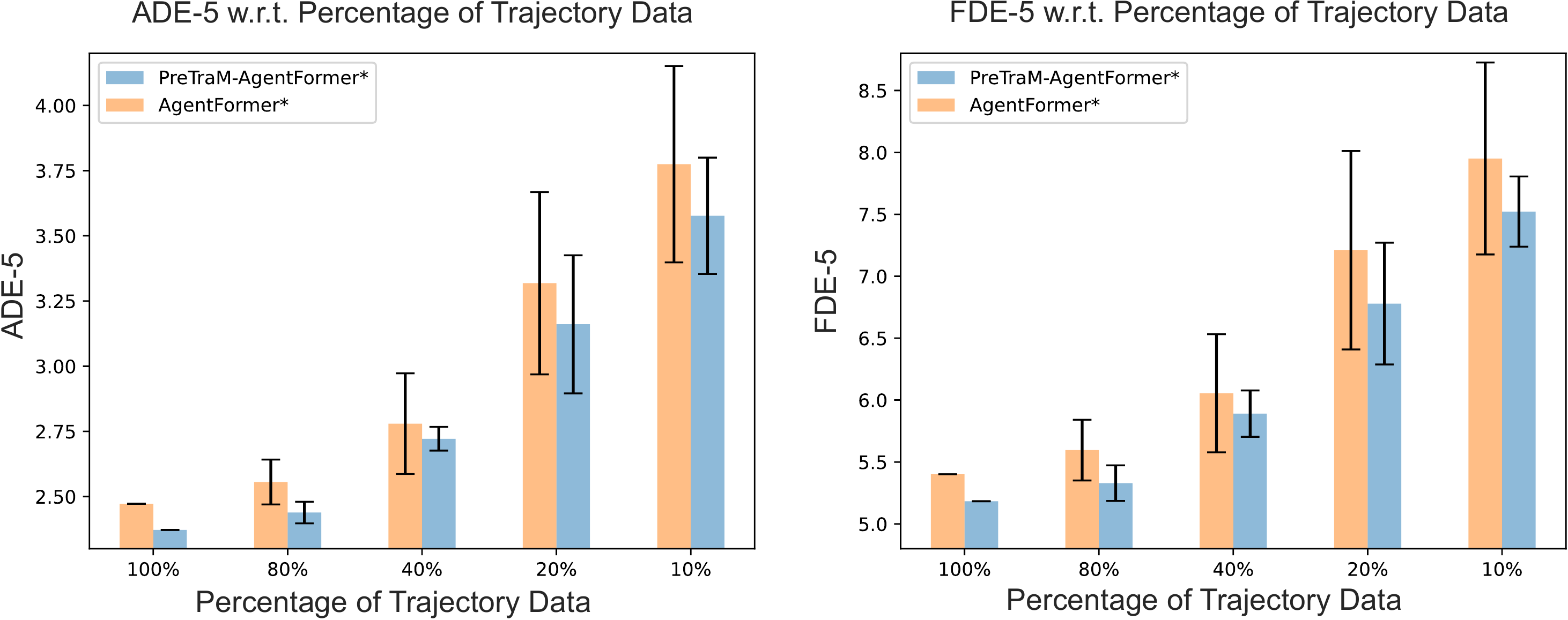}
	\end{center}
\caption{Experiments with part of the trajectory data. Left: ADE-5 results. Right: FDE-5 results. We repeat the experiments with 3 different random seeds and report the mean performance. The error bars are 3 times the standard deviation. As the percentage of trajectory data becomes lower, the improvements of PreTraM are larger. Moreover, the std of PreTraM is much smaller than the baseline over all the settings.}
\label{fig:dataeff}
\end{figure}

\subsection{Data Efficiency}
\label{sec:data-eff}
In this section, we explore whether the learned representations of trajectory and map can improve data efficiency. To investigate this  we evaluate PreTraM-AgentFormer on a fraction of the dataset, comparing its result with baseline AgentFormer.  In this set of experiments to best demonstrate our strength, we use ResNet18 as a substitute for the 4-layer map encoder, Map-CNN, in the original AgentFormer.
This is due to the intuition that larger models are better at representation learning \cite{chen2020simple,radford2021learning,he2020momentum}.
We randomly sample 80\%, 40\%, 20\% and 10\% trajectories from the dataset, but keep all the HD-maps available. For each setting, we repeat the experiments with 3 different random seeds and report the mean and the standard deviation in Figure \ref{fig:dataeff}.
We observe that PreTraM-AgentFormer outperforms the baseline in all settings. More importantly, the performance gain of PreTraM gets larger as the percentage of data goes smaller. With \textit{10\%} of data, i.e., around 1200 samples, PreTraM surpasses the baseline by \textit{0.32} on ADE-5 and \textit{0.77} on FDE-5. Moreover, the std of PreTraM is much smaller than the baseline. The std of ADE-5 of the baseline are 0.035 (80\%), 0.079 (40\%), 0.143 (20\%), and 0.154 (10\%) respectively, while those of PreTraM are 0.017, 0.018, 0.108, 0.091. It is the same case in terms of FDE-5.

In addition, we observe that training on 70\% of data, PreTraM-AgentFormer still outperforms the baseline with 100 \% of data (2.470 ADE-5 vs. 2.472 ADE-5). More results are shown in supplementary material.

\begin{figure}[!t]
	\begin{center}
		\includegraphics[width=\linewidth]{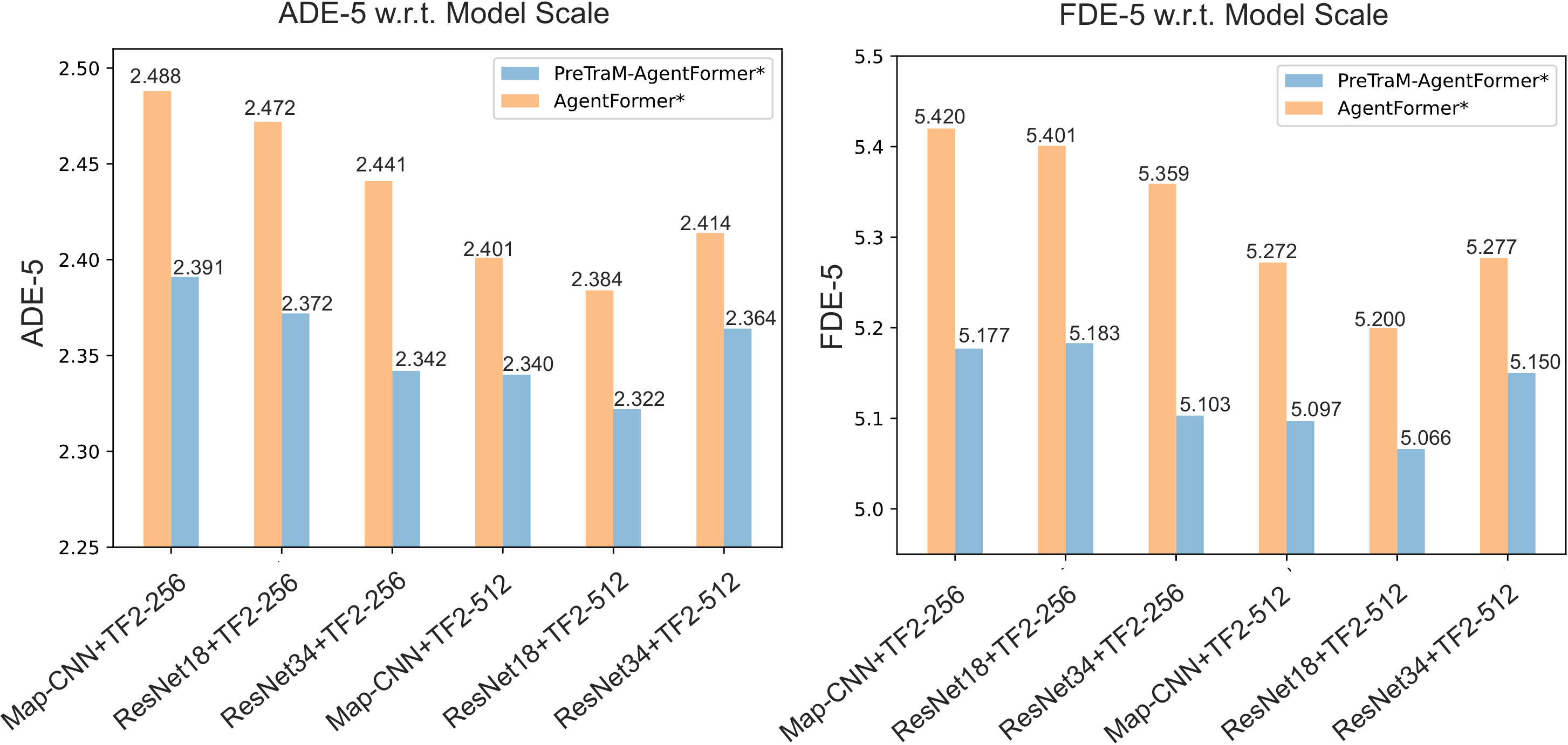}
	\end{center}
	\caption{Experiment with models at different scales. As the model gets deeper and wider, PreTraM consistently improves AgentFormer by a large margin.}
	\label{fig:modelscale}
\end{figure}



\subsection{Scalability Analysis}

A good representation learning method is able to scale with the model size~\cite{he2020momentum,chen2020simple}. Therefore, we evaluate PreTraM on map encoders and trajectory encoders of different depth and width. 
 Map-CNN is the map encoder used in original AgentFormer. It is merely a 4-layer convolutional network. Alternatively, we substitute ResNet18 and ResNet34 as the map encoder. Besides, we also increase the width of the trajectory encoder, a 2-layer Transformer encoder~\cite{vaswani2017attention}, from 256-channel to 512-channel.
 As shown in Figure~\ref{fig:modelscale}, PreTraM consistently improves ADE and FDE upon models of different scales by a large margin.

\subsection{Analysis} \label{sec:analysis}
It is natural that PreTraM enhances the map representation since we utilize 28.8M maps for pre-training the map encoder, but as we proposed in previous sections, another important goal is to further enhance the trajectory representation. Therefore, we conduct experiments and delve deep into the function of PreTraM to discuss how it enhances the trajectory representation. All the experiments are completed upon AgentFormer with Map-CNN.

\textbf{Does PreTraM indeed improve trajectory representation?}
In fact, we can quantitatively demonstrate this by loading one of the pre-trained encoders, trajectory encoder (TE) and map encoder (ME), at finetuning phase. As shown in Table \ref{tab:ablation_finetuning}, we can first observe that loading only the pre-trained map encoder improves prediction performance. More importantly, we observe that just loading TE pre-trained weights is able to give almost the same result as loading both of ME and TE. This means the learnt trajectory representation is strong, and that the major benefit of PreTraM owes to the trajectory representation.

So our answer is ``Yes, PreTraM indeed improves trajectory representation."


\begin{table}[!t]
\centering

\caption{Comparison with loading one of the pretrained models when finetuning. ME: map encoder, TE: past trajectory encoder.}
\begin{tabular}{ c c c  }
 \hline
Method & ADE-5 & FDE-5\\
 \hline
 Baseline AgentFormer & 2.488 & 5.420\\
 \hline
Finetune with both pretrained weights (PreTraM) &2.391\color{ForestGreen}{(-0.097)} & 5.177\color{ForestGreen}{(-0.243)} \\
Finetune with only TE pretrained weights & 2.399\color{ForestGreen}{(-0.089)}&5.277\color{ForestGreen}{(-0.143)} \\
Finetune with only ME pretrained weighst & 2.454\color{ForestGreen}{(-0.034)}& 5.372\color{ForestGreen}{(-0.048)} \\

\hline
\end{tabular}
\label{tab:ablation_finetuning}
\end{table}
\begin{table}[!t]
\centering
\caption{Comparison with different pre-training strategies. MTM means masked trajectory modeling, recovering the masked trajectories during pre-training, which is a mimic of masked language modeling in NLP \cite{devlin2019bert}. }
\begin{tabular}{ c c c  }
 \hline
Method & ADE-5 & FDE-5\\
 \hline
Baseline AgentFormer & 2.488 & 5.420\\
\hline
Pre-training with both TMCL and MCL (PreTraM) &2.391\color{ForestGreen}{(-0.097)} & 5.177\color{ForestGreen}{(-0.243)} \\
Pre-training with MTM and MCL &2.431\color{ForestGreen}{(-0.057)} & 5.322\color{ForestGreen}{(-0.098)} \\
Pre-training with only MCL & 2.442\color{ForestGreen}{(-0.046)} &5.373\color{ForestGreen}{(-0.057)} \\
Pre-training with only TMCL & 2.451\color{ForestGreen}{(-0.037)} & 5.369\color{ForestGreen}{(-0.051)} \\
\hline
\end{tabular}
\label{tab:ablation_pretrain}
\end{table}

\textbf{Is TMCL crucial for improving trajectory representation?} To examine the contribution of TMCL, we experiment with an alternative to TMCL as the objective function for pre-training trajectory representation. Inspired by Masked Language Modeling (MLM)~\cite{devlin2019bert} for sequence modeling in NLP, we randomly mask out part of the input history states and ask the trajectory encoder to recover the masked part. Denoting this task as Masked Trajectory Modeling (MTM), we jointly pre-train the model on the objective of MTM and MCL.
For variable controlling, we also pre-train the model solely on MCL.
As shown in Table \ref{tab:ablation_pretrain} we find that MTM plus MCL does improve from the baseline but is almost comparable to pre-training with only MCL. It shows the important role of TMCL in trajectory representation learning as it learns trajectory-map relationship and bridges the trajectory and map embedding space.

So our answer is ``Yes, TMCL is crucial to improve trajectory representation."




\textbf{Is MCL crucial for improving trajectory representation?}
Indeed, as shown in Table \ref{tab:ablation_pretrain}, when pre-training only with MCL, the improvement is 0.046. This makes sense in that HD-map is an important prior to prediction and thus better map representation in general can improve prediction. But is MCL also helpful to trajectory representation? To examine this, we only pre-train with TMCL. We find that without MCL, TMCL brings limited improvements compared with PreTraM, \textit{e.g.,} 0.037 ADE-5 vs. 0.097 ADE-5 with PreTraM (Table. \ref{tab:ablation_pretrain}). This demonstrates that although map and trajectory are totally different modalities, PreTraM makes use of much more maps to enhance trajectory representation under the situation that the trajectory data is limited. 

So our answer is ``Yes, MCL is crucial to improve trajectory representation."

\section{Related works}
In this work, we focus on the pre-training of trajectory and map embeddings for trajectory forecasting. Given a trajectory prediction model, the applicable pre-training schemes largely depend on the adopted scene representation. In this section, we first give a concise summary of the trajectory prediction literature from the perspective of scene representation. Then, we review several works related to pre-training for trajectory forecasting. Since pre-training is under-explored in the field, we broaden the discussion to self-supervised learning. 
\subsection{Scene Representation in Trajectory Forecasting} \label{sec:review-representation}
In complex urban traffic scenarios, it is crucial to utilize the semantic information of the scene, including the map and surrounding agents, to make accurate predictions. A widely-adopted approach is to employ rasterized top-down semantic images around the target agents as input and use CNNs to encode the context \cite{covernet,chai2019multipath,cui2019multimodal,salzmann2020trajectron++,yuan2021agent,deo2022multimodal}. The history trajectories of the predicted agents are encoded separately and then aggregated with the context embedding. Our proposed PreTraM can be directly applied to pretrain models adopting this rasterized scene representation. The image-based representation has constant input size regardless of the complexity of the scene, which makes encoding simple and unified. However, some argued recently that rich semantic and structured information (e.g., relations between road segments) of the maps is lost through rasterization \cite{vectornet,liang2020learning}. To this end, they proposed to represent the scenes as graphs that naturally inherit the maps' structured information. Graph neural networks (GNN) \cite{battaglia2018relational,liang2020learning} and Transformers \cite{vaswani2017attention,vectornet} were then adopted to encode the context information from the scene graphs. Following their works, many graph-based models have achieved state-of-the-art performance on multiple prediction benchmarks \cite{zhao2020tnt,gilles2021gohome,gu2021densetnt,gilles2021thomas,varadarajan2021multipath++,deo2022multimodal}.

\subsection{Self-Supervised Learning in Trajectory Forecasting}

Pre-training and, in a broader sense, self-supervised learning are under-explored for trajectory forecasting. There are only a few recent works investigating their applications in trajectory prediction. Inspired by similar methods in NLP, an auxiliary graph completion task was proposed in \cite{vectornet} to enhance the node representation, including both road elements and agents. However, the graph completion objective was jointly optimized with the prediction task. Moreover, the auxiliary task was applied to their Transformer-based encoder for the scene graphs, which limits the amount of data for self-supervised training to the size of prediction datasets. In contrast, our PreTraM framework lets trajectory encoder benefit from the large number of map patches that are not associated with agents. In \cite{hengbo2021selfsupervised}, SimCLR was adopted to pre-train the representation of rasterized maps and agent relations. They deliberately introduced assumptions on semantic invariant operations based on domain knowledge. In our MCL, we follow \cite{gao2021simcse} to avoid any assumptions on semantic invariant operations. Moreover, \cite{hengbo2021selfsupervised} focuses on contrastive learning within the same modality of data, whereas our PreTraM framework leverages both single-modal and cross-modal contrastive learning to jointly train the trajectory and map representations. As shown in Sec. \ref{sec:analysis}, PreTraM has clear performance advantage over single-modal contrastive learning within each data modality. The trajectory representation indeed benefits from the shared map-trajectory embedding space induced by TMCL. 

\section{Discussion and Limitations}
In our experiments, we demonstrated that PreTraM is effective for prediction models based on rasterized map representation. As reviewed in Sec. \ref{sec:review-representation}, graph-based methods have become more popular due to their ability in preserving the rich semantic and structured information of HD-maps. In principle, PreTraM is not limited to image-based map encoders. As along as we can obtain separate map and trajectory embeddings from the prediction pipeline, we can apply PreTraM to those graph-based models. For instance, some works applied a two-stage graph encoding scheme, where the map graph was encoded before being fused with trajectory embeddings \cite{liang2020learning,gilles2021gohome,gilles2021thomas}. However, GNNs may behave differently from CNNs during pre-training. We are interested in extending PreTraM to graph-based scene representation and evaluating its performance in our future study. Meanwhile, in other works, the road elements and agents were integrated into a single graph before aggregation \cite{vectornet,zhao2020tnt,gu2021densetnt,varadarajan2021multipath++,deo2022multimodal}. PreTraM cannot be applied to these models, as there are no matching pairs of map and trajectory embeddings in their pipelines. We are also interested in exploring alternative pre-training methods for those models in the future.

\section{Conclusion}
In this paper, we propose PreTraM, a novel self-supervised pre-training scheme for trajectory prediction. We design Trajectory-Map Contrastive Learning (TMCL) to help models capture the relationship between agents and the surrounding HD-map, and Map Contrastive Learning (MCL) to enhance map representation via a large number of augmented map patches that are not associated with the agents. With PreTraM, we enhance the prediction performance of Trajectron++ and AgentFormer by 5.5\% and 6.9\% relatively on FDE-10. Furthermore, PreTraM can improve the data efficiency of the prediction models. We also demonstrate that our method can consistently improve performance when the model size scales up. Through ablation studies and analysis, we show that PreTraM indeed enhances map and trajectory representations. In particular, a better trajectory representation is learned via bridging the map and trajectory representations with TMCL, so that the trajectory encoder can benefit from the map representation enhanced by MCL. It shows that the performance improvement is attributed to the coherent integration of MCL and TMCL in our framework. 


%
%
\bibliographystyle{splncs04}
\bibliography{egbib}



\clearpage
\appendix

\section{Data Efficiency}
In section \ref{sec:data-eff}, we demonstrate with experiments that PreTraM boosts data efficiency of the prediction model. Apart from the results shown there, we perform experiments at other percentages of trajectory data, ranging from 80\%, 70\%, 60\% to 20\% and 10\%. Besides, to ensure reliable results for lower percentages including 20\% and 10\%, we repeat experiments for 5 and 10 times respectively. For the other settings, we repeat 3 times as in section \ref{sec:data-eff}. We report the mean performance as well as the variance in table \ref{tab:data-efficiency}. Note that these experiments are all based on AgentFormer with ResNet18.

On this full set of experiment results, we observe that across all settings of experiments, PreTraM mitigates prediction error and significantly reduces variance of performance. For the mean value of using different random seeds with using 70\% data, PreTraM-AgentFormer gives 2.451 ADE-5 and 5.343 FDE-5, which outperforms the baseline with 100\% data. Note that for one specific experiment with 10 \% data, we observe that PreTraM-Agentformer can achieve 0.32 ADE-5 and 0.77 FDE-5 improvement, as reported in the main text.

Therefore, we conclude that PreTraM indeed enhances data efficiency.

\begin{table}[b]
\centering

\caption{Experiment results with part of the trajectory data. For percentages above 30\%, experiments are repeated 3 times, while for 20\% and 10\% data, the experiments are repeated 5 and 10 times respectively.}
\label{tab:data-efficiency}
\begin{tabular}{ c| c| c| c | c}
 \hline
\multirow{2}{6em}{Percentage of Trajectory }& \multicolumn{2}{c|}{AgentFormer} & \multicolumn{2}{c}{PreTraM-AgentFormer}\\
\cline{2-5}
 & ADE-5 & FDE-5 & ADE-5 & FDE-5 \\
 \hline
100\% & 2.472 & 5.401 & 2.372 & 5.183  \\
80\%  & 2.556$\pm$0.035 & 5.599$\pm$0.071 & 2.439$\pm$0.017 & 5.329$\pm$0.059  \\
70\% & 2.570$\pm$0.038 & 5.609$\pm$0.075 & 2.451$\pm$0.030 &    5.343$\pm$0.072\\
60\%  & 2.586$\pm$0.051 & 5.621$\pm$0.123 & 2.524$\pm$0.031 & 5.477$\pm$0.103 \\
50\% & 2.707$\pm$0.037 & 5.889$\pm$0.089 & 2.599$\pm$0.016 & 5.652$\pm$0.030    \\
40\%  &  2.780$\pm$0.079 & 6.055$\pm$0.195 & 2.721$\pm$0.018 & 5.888$\pm$0.081 \\
30\% & 3.055$\pm$0.061 & 6.672$\pm$0.101 & 2.890$\pm$0.065  & 6.281$\pm$0.140  \\
20\%  & 3.388$\pm$0.115 & 7.339$\pm$0.256 & 3.180$\pm$0.089 & 6.840$\pm$0.181 \\
10\% & 3.818$\pm$0.141  & 8.038$\pm$0.292 & 3.618$\pm$0.114 &  7.607$\pm$0.229   \\
\hline

\end{tabular}
\end{table}

\section{Ablation Study on Augmentation Operation in MCL}

MCL uses dropout as minimal augmentation. In section \ref{sec:MCL}, we briefly discuss why dropout works better for map representation learning (MCL) than the commonly used data augmentation applied on contrastive learning in computer vision.

To verify the argument, we conduct experiments with different augmentation operations including rotation, flip, color jitter, and gaussian noise \cite{chen2020simple}. The experiments are based on PreTraM-AgentFormer with ResNet18, and the results are reported in Table \ref{tab:aug}.
We observe that MCL via any of the alternative data augmentations even deteriorates the performance of AgentFormer, rather than boosting it. 
They are worse than the baseline AgentFormer by at least 0.019 ADE-5 and 0.076 FDE-5. In contrast, PreTraM with MCL via dropout drastically improves the baseline by 0.1 ADE-5 and 0.218 FDE-5. This sharp contrast indicates that conventional augmentations do destroy the semantic and the topology in the HD-map, while dropout acting as minimal augmentation enhances the representation learning of HD-maps and consequently promotes TMCL.

\begin{table}[!t]
\centering
\caption{We use different augmentation operations to conduct MCL for comparison. MCL via dropout is our proposed method.}
\label{tab:aug}
\begin{tabular}{ c c c  }
 \hline
PreTraM-AgentFormer with different MCL & ADE-5 & FDE-5\\
\hline
AgentFormer (baseline)& 2.472& 5.401 \\
PreTraM-AgentFormer (MCL via dropout) & 2.372\color{ForestGreen}{(-0.100)} & 5.183\color{ForestGreen}{(-0.218)}\\
 \hline
PreTraM-AgentFormer (MCL via random rotation) & 2.508\color{red}{(+0.036)}&5.477\color{red}{(+0.076)} \\
PreTraM-AgentFormer (MCL via random flip) &2.491\color{red}{(+0.019)} & 5.499\color{red}{(+0.098)} \\
PreTraM-AgentFormer (MCL via color jitter) & 2.505\color{red}{(+0.033)}&5.505\color{red}{(+0.104)} \\
PreTraM-AgentFormer (MCL via gaussian noise) &2.541\color{red}{(+0.069)} &  5.601\color{red}{(+0.200)}\\

\hline
\end{tabular}
\end{table}

\section{Implementation Details}
All of our experiments are developed based on Pytorch and conducted on Intel Haswell CPU platform with one single V100 GPU.

\textbf{Reproduced AgentFormer.} We refer to the official code of AgentFormer to reproduce a new version that supports parallel computing. Specifically, original AgentFormer only uses one single scene in each iteration due to the variant agent number in different scenes. Therefore, to support parallel computing, we pad the agent number to 20, and the padding agents will be masked in the models as well as during loss computation so that the whole process is not influenced by the padding agents. On the other hand, we do not use the step decay learning rate scheduler in the original AgentFormer, but use the linear scheduler with warmup provided by HuggingFace~\cite{wolf-etal-2020-transformers}, which is commonly used in language-targeted transformer models \cite{devlin2019bert}. During training, we set the batch size as 8, the initial learning rate as $10^{-4}$, the warmup rate as 0.1, and we train the model with Adam optimizer for 100 epochs. During inference, we keep the same setting as the original AgentFormer.

\textbf{PreTraM on top of AgentFormer and Trajectron++.} We simply conduct PreTraM on map encoder and past encoder of AgentFormer and Trajectron++ in the pre-training phase. Slightly different from AgentFormer, the past feature encoding in Trajectron++ includes past trajectory encoding and edge encoding, we concatenate them as the trajectory features in PreTraM. 

We avoid taking great pains to tune hyperparameters of the pre-training phase and use the same training recipe as the finetuning phase. The only difference is that we pretrain the models with fewer epochs (20 epochs for AgentFormer and 5 epochs for Trajectron++). Besides, we set $\lambda$ in equation (4) to 1. These choices, though simple, turn out to work well for PreTraM.


We provide a numpy-like pseudocode of PreTraM, as shown below. 

\begin{lstlisting}[language=Python, caption=Numpy-like pseudocode of PreTraM.]
# past_trajectory_encoder: Transformer Encoder in AgentFormer or LSTM in Trajectron++
# map_encoder: Map_CNN or ResNet with dropout
# s[N_traj, T, D]: input trajectories
# m[N_traj, C, C, 3]: agent-centric map patches paired with s
# m_mcl[N_map, C, C, 3]: trajectory-decoupled maps
# W_traj[d_s, d_e]: learned projection of trajectory to embed
# W_map[d_m, d_e]: learned projection of map to embed (TMCL)
# W_mcl[d_m, d_e]: learned projection of map to embed (MCL)
# t_traj: learned temperature parameter for TMCL
# t_map: learned temperature parameter for MCL
# lamda: hyperparameter for balancing MCL and TMCL

####################
## For TMCL
####################
# extract feature
h_traj_timed = past_trajectory_encoder(s) # [N_traj, T, d_t]
h_map = map_encoder(m) # [N_traj, d_m]

# mean pooling over time dimension
h_traj = h_traj_timed.mean(axis=1) # [N_traj, d_t]

# linear projection to embedding space [N_traj, d_e]
h_traj_norm = l2_normalize(np.dot(h_traj, W_traj), axis=1)
h_map_norm = l2_normalize(np.dot(h_map, W_map), axis=1)

# compute pairwise similarities [N_traj, N_traj]
logits_tmcl = np.dot(h_traj_norm, h_map_norm.T) / t_traj

# symmetric loss
labels = np.arange(N_traj)
loss_traj = cross_entropy_loss(logits_tmcl, labels)
loss_map = cross_entropy_loss(logits_tmcl.T, labels)
loss_TMCL = (loss_traj + loss_map)/2
####################
## For MCL
####################
# extract feature [N_map, d_m]
h_dp1 = map_encoder(m_mcl)  
h_dp2 = map_encoder(m_mcl) # different dropout mask

# shared linear projection [N_map, d_e]
h_dp1_norm = l2_normalize(np.dot(h_dp1, W_mcl), axis=1)
h_dp2_norm = l2_normalize(np.dot(h_dp2, W_mcl), axis=1)

# compute pairwise similarities [N_map, N_map]
logits_mcl = np.dot(h_dp1_norm, h_dp2_norm.T) / t_map

# contrastive loss
labels = np.arange(N_map)
loss_MCL = cross_entropy_loss(logits_mcl, labels)

# Total loss
loss_total = loss_TMCL + lamda * loss_MCL


\end{lstlisting}


\end{document}